\documentclass[letterpaper]{article} % DO NOT CHANGE THIS
\usepackage{aaai25}  % DO NOT CHANGE THIS
\usepackage{times}  % DO NOT CHANGE THIS
\usepackage{helvet}  % DO NOT CHANGE THIS
\usepackage{courier}  % DO NOT CHANGE THIS
\usepackage[hyphens]{url}  % DO NOT CHANGE THIS
\usepackage{graphicx} % DO NOT CHANGE THIS
\usepackage{natbib}  % DO NOT CHANGE THIS AND DO NOT ADD ANY OPTIONS TO IT
\usepackage{caption} % DO NOT CHANGE THIS AND DO NOT ADD ANY OPTIONS TO IT
\usepackage{algorithm}
\usepackage{algorithmic}

\usepackage{newfloat}
\usepackage{amssymb}
\usepackage{amsmath}
\usepackage{listings}
\usepackage{multirow}
\usepackage{booktabs}
\usepackage{subcaption}
\usepackage{upgreek}
\usepackage[table]{xcolor}
\usepackage{url}

\DeclareCaptionStyle{ruled}{labelfont=normalfont,labelsep=colon,strut=off} % DO NOT CHANGE THIS
\floatstyle{ruled}
\newfloat{listing}{tb}{lst}{}
\floatname{listing}{Listing}
\newcommand{\citepyear}[1]{\cite{#1}}

\title{Pre-training a Density-Aware Pose Transformer for Robust LiDAR-based 3D Human Pose Estimation}
\author{
    Xiaoqi An\textsuperscript{\rm 1},
    Lin Zhao\textsuperscript{\rm 1}\footnote{Corresponding authors.},
    Chen Gong\textsuperscript{\rm 1},
    Jun Li\textsuperscript{\rm 1},
    Jian Yang\textsuperscript{\rm 1}\footnotemark[1]
}

\affiliations{
    \textsuperscript{\rm 1}PCA Lab, Key Lab of Intelligent Perception and Systems for High-Dimensional Information of Ministry of Education\\
    School of Computer Science and Engineering, Nanjing University of Science and Technology\\
    \{xiaoqi.an, linzhao, chen.gong, junli, csjyang\}@njust.edu.cn,
}

\usepackage{bibentry}
\begin{document}

\maketitle

\begin{abstract}
With the rapid development of autonomous driving, LiDAR-based 3D Human Pose Estimation (3D HPE) is becoming a research focus. However, due to the noise and sparsity of LiDAR-captured point clouds, robust human pose estimation remains challenging. Most of the existing methods use temporal information, multi-modal fusion, or SMPL optimization to correct biased results. In this work, we try to obtain sufficient information for 3D HPE only by modeling the intrinsic properties of low-quality point clouds. Hence, a simple yet powerful method is proposed, which provides insights both on modeling and augmentation of point clouds.
Specifically, we first propose a concise and effective density-aware pose transformer (DAPT) to get stable keypoint representations. By using a set of joint anchors and a carefully designed exchange module, valid information is extracted from point clouds with different densities. Then 1D heatmaps are utilized to represent the precise locations of the keypoints. 
Secondly, a comprehensive LiDAR human synthesis and augmentation method is proposed to pre-train the model, enabling it to acquire a better human body prior. We increase the diversity of point clouds by randomly sampling human positions and orientations and by simulating occlusions through the addition of laser-level masks.
Extensive experiments have been conducted on multiple datasets, including IMU-annotated LidarHuman26M, SLOPER4D, and manually annotated Waymo Open Dataset v2.0 (Waymo), HumanM3. Our method demonstrates SOTA performance in all scenarios. In particular, compared with LPFormer on Waymo, we reduce the average MPJPE by $10.0mm$. Compared with PRN on SLOPER4D, we notably reduce the average MPJPE by $20.7mm$. 
\end{abstract}

% Uncomment the following to link to your code, datasets, an extitended version or similar.
%
\begin{links}
    \link{Code}{https://github.com/AnxQ/dapt}
\end{links}

\section{Introduction}
3D human pose estimation (3D HPE) is a fundamental computer vision task with a wide range of downstream usages such as human behavior understanding{}, trajectory prediction{}, autonomous driving \cite{congSTCrowdMultimodalDataset2022,lianTransformerReID2022}, etc. To implement 3D HPE, a simple and direct way is to regress 3D keypoint coordinates directly from 2D HPE results \cite{kang2023gridconv, gongPoseAugDifferentiablePose2021a}. However, these methods have difficulty predicting world coordinates, and robust pose estimation in the real scenario remains challenging.

To get global 3D keypoints, most traditional 3D HPE methods are based on multi-view RGB images \cite{simonHandKeypointDetection2017,iskakovLearnableTriangulationHuman2019,tuVoxelPoseMulticamera3D2020,yeFasterVoxelPoseRealtime2022a} or RGB-D images \cite{yingRgbDFusionPointCloudBased2021a,hongDynamicPoseEstimation2018}, which require in-door laboratory environments with complex calibrations \cite{suRobustFusionHumanVolumetric2020,zhengMultimodal3DHuman2022}. Alternatively, LiDAR sensors can obtain accurate point-level depth in complex environments, which are more adaptable to long-range 3D HPE in open-world scenes.

\begin{figure}[t]
  \centering
  \includegraphics[width=1.0\linewidth]{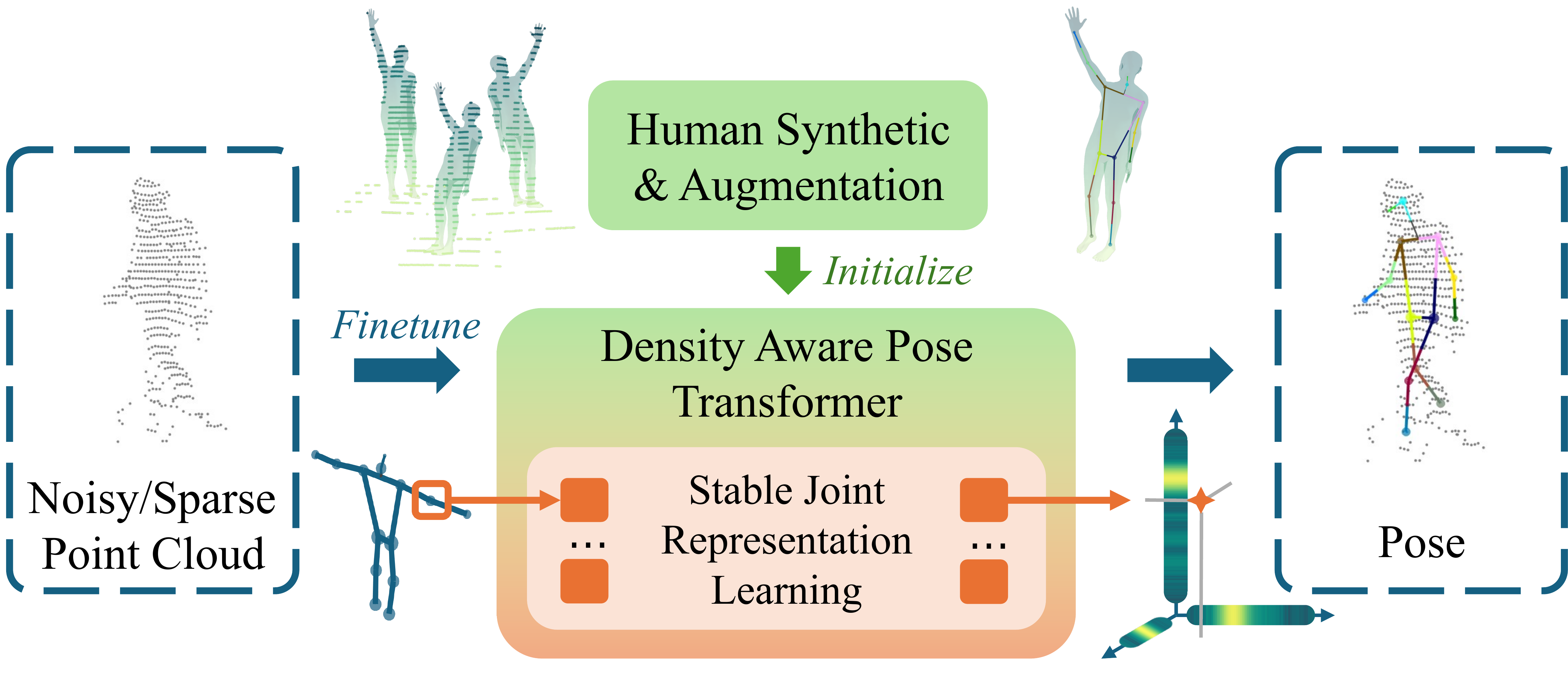}
  \caption{We propose a novel framework for LiDAR-based 3D HPE. With comprehensive LiDAR human synthesis \& augmentation for model pre-training as well as the learning of stable joint representations, our method produces robust results from low-quality point clouds.}
  \label{fig:intro}
\end{figure}

\begin{figure}[t]
  \centering
  \includegraphics[width=1.0\linewidth]{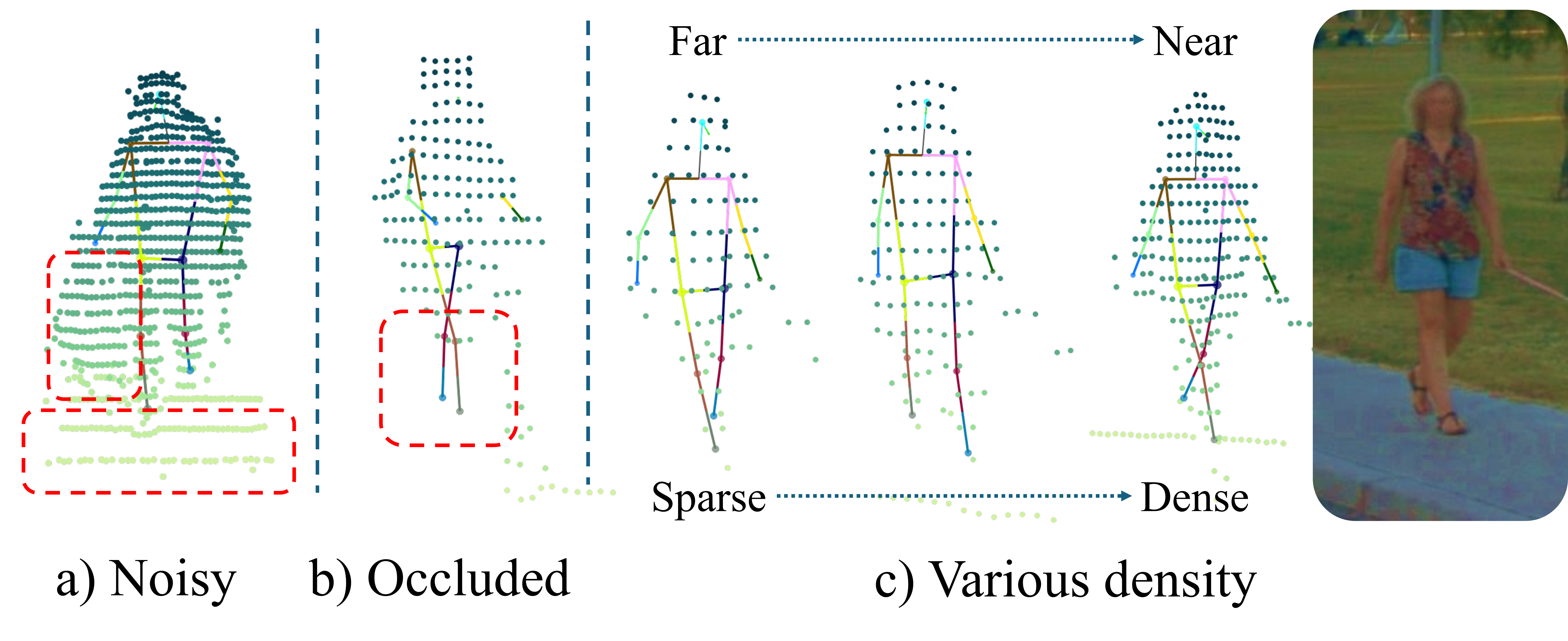}
  \caption{Difficulties of LiDAR-based 3D HPE, mainly lie in samples with noisy, occluded, or sparse point clouds.}
  \label{fig:lidar_difficult}
\end{figure}

However, compared to clear and dense point clouds generated by RGB-D images \cite{fanSURREALOpenSourceReinforcement2018,ionescuHuman36MLarge2014a}, the LiDAR-captured point clouds have various inconsistencies, making them difficult to learn directly with existing methods. As shown in Fig.\ref{fig:lidar_difficult}, the samples in Waymo Open Dataset \cite{wod2020cvpr} have various point densities, and the noisy points from the environment may lead to ambiguity and unstable predictions. Therefore, to achieve robust pose estimation, \cite{liLiDARCapLongrangeMarkerless2022,yanRELI11DComprehensiveMultimodal2024,zhao2024graph} integrate ST-GCNs to get stabilized pose results from multiple LiDAR frames. \cite{congWeaklySupervised3D2022,furstHPERL3DHuman2021} perform multi-modal fusion to mitigate the lack of information in sparse point clouds, and \citepyear{liLiDARCapLongrangeMarkerless2022,yanRELI11DComprehensiveMultimodal2024,zhangNeighborhoodEnhanced3DHuman2024} introduce SMPL \cite{smpl2015} as a priori to further align human mesh with point clouds. Despite the success of these methods, they inevitably introduce additional data acquisition or time-consuming optimization processes which complicate the entire framework and make it less suitable for practical applications.

In this paper, we propose a novel framework to learn stable representations for 3D HPE only using single-frame low-quality LiDAR point clouds. Our method is simple yet effective. Specifically, We provide insights into both model design for accurate pose estimation and model pre-training for effective body prior learning.

\begin{figure}[t]
  \centering
  \includegraphics[width=0.7\linewidth]{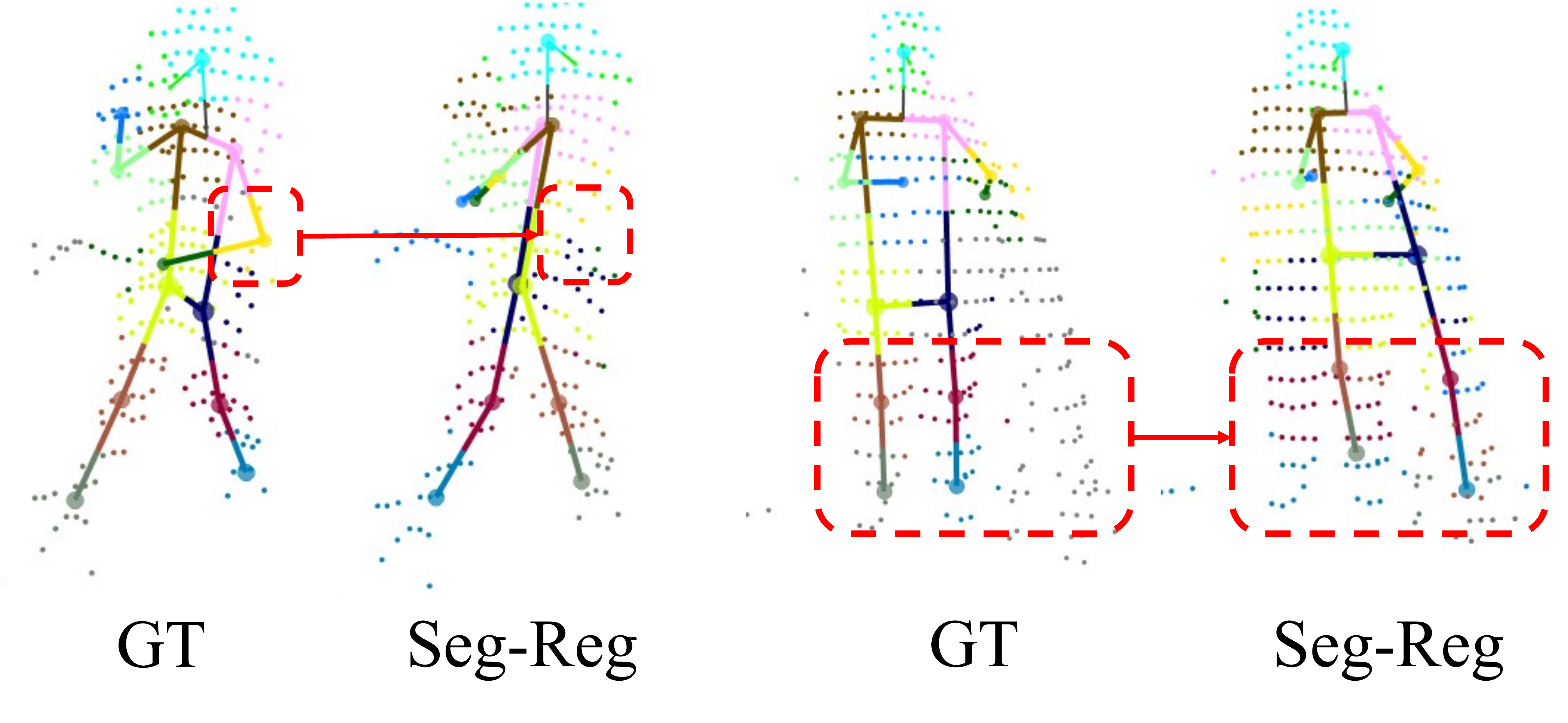}
  \caption{Results of the segment-regression-based methods. The color of the point indicates which joint it belongs to. The wrong segmentation leads to a biased joint location.}
  \label{fig:false_seg}
\end{figure}

Firstly, we design a \textbf{D}ensity-\textbf{A}ware \textbf{P}ose \textbf{T}ransformer (DAPT) that provides stable and explicit representations of joints. Most existing models attempt to regress the coordinates of joints from global or clustered point-wise features. This makes the model highly susceptible to the density and spatial distribution of the point cloud. As shown in Fig.\ref{fig:false_seg}, when the point cloud near the joint is sparse or noisy, the network cannot correctly identify which body part the point belongs to, resulting in a biased joint location. To solve this problem, we introduce a set of learnable joint anchors. When extracting point cloud features, they can explicitly integrate the information across multiple density levels through a carefully designed exchange module. Then, we utilize 1D-heatmaps on the XYZ axes to represent joint positions, which allows the model to obtain a stable output.

Secondly, we introduce a comprehensive pre-training approach that conducts thorough LiDAR human synthesis and augmentation. Since annotating LiDAR data is expensive, inspired by \citepyear{weng3DHumanKeypoints2023}, we train on samples synthesized by SMPL mesh under ray casting. To tackle noisy and sparse point clouds in real scenarios, we add a square surface that can be varied in small magnitudes, such as the ground surface, and put the human mesh into the scene with randomized poses and positions. On the other hand, since the light beams are highly susceptible to occlusion by foreground objects, we mask the range image with patches to simulate the occlusion. Therefore, the model will learn prior knowledge of the human body and mine important clues about the pose in low-quality point clouds.

With the synergy of pre-training and DAPT, our approach provides a robust way to understand human body configurations in outdoor environments. Although the proposed method is optimization-free and does not use information from any other modalities or time series, we still outperform SOTA methods on multiple datasets. Specifically, our method compared with LPFormer \citepyear{yeLPFormerLiDARPose2023} on the manually annotated Waymo Open Dataset \cite{wod2020cvpr} reduces the mean per joint position error (MPJPE) by 10mm ($\downarrow$16\%). When compared with PRN \cite{fanLiDARHMR3DHuman2023} on the IMU-annotated SLOPER4D dataset \cite{daiSLOPER4DSceneAwareDataset2023}, it reduces the MPJPE by 20.7mm ($\downarrow$58\%).

In summary, our contribution lies in three main aspects:
\begin{itemize}
    \item We propose a density-aware pose transformer that steadily mines pose cues from sparse and noisy point clouds.
    \item We thoroughly investigate the difficulties of estimating human pose by LiDAR data and design a comprehensive pre-training approach.
    \item Our method greatly improves the stability and accuracy of single-frame LiDAR-only human pose estimation, achieving SOTA performance in multiple scenarios.
\end{itemize}

\section{Related works}
% \subsection{Optical-based 3D human pose estimation}
\subsection{LiDAR-based 3D human pose estimation}
In recent years, many LiDAR point cloud-based 3D HPE methods have been proposed as the practical application value of LiDAR has been explored. \cite{liLiDARCapLongrangeMarkerless2022} provides the first LiDAR HPE dataset and proposes the first fully supervised baseline for LiDAR-based motion capture. The keypoint coordinates are obtained by a temporal encoder and optimized with inverse kinematics and SMPL. \cite{yanCIMI4DLargeMultimodal2023} provides a climbing dataset with explicit scene interactions and attempts to perform scene-aware human pose estimation, followed by \cite{zhangNeighborhoodEnhanced3DHuman2024} which uses environmental information of 3D neighbors sampled in the background to enhance the pose learning. \cite{renLiveHPSLiDARbasedScenelevel2024a,renLiveHPSRobustCoherent2024} achieves accurate motion tracking by exploiting temporal and spatial coherence. On the other hand, \cite{congWeaklySupervised3D2022,zhengMultimodal3DHuman2022,furstHPERL3DHuman2021,hu2024towards} perform multi-modal fusion, which utilizes the keypoint cues and geometric constraints provided by the 2D images for weakly-supervised 3D human pose learning. \cite{yeLPFormerLiDARPose2023} proposes a multitasking architecture that augments model learning with segmentation and object detection tasks, and uses a keypoint transformer for multi-person 3D HPE.

\begin{figure*}[t]
  \centering
  \includegraphics[width=0.95\textwidth]{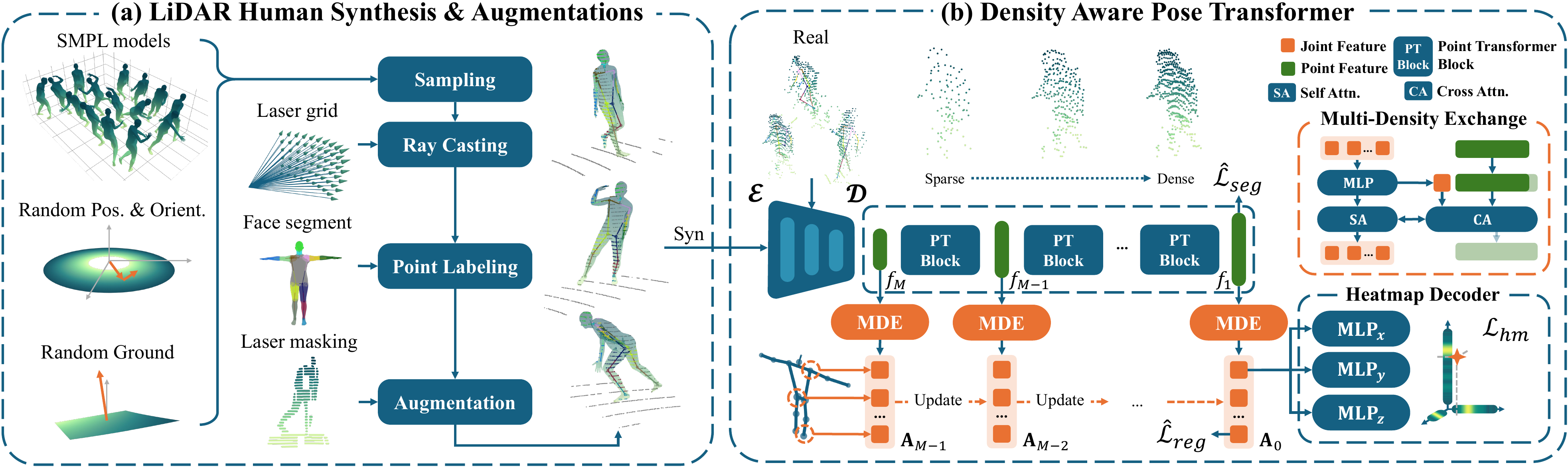}
  \caption{Overall structure of our method. It mainly consists of \textbf{(a)} a comprehensive LiDAR human synthesis and augmentation framework to provide internal human priors, and \textbf{(b)} a Density Aware Pose Transformer that uses the multi-density exchange (MDE) module to extract stable joint representations from point cloud features.}
  \label{fig:framework}
\end{figure*}

Despite the success of these approaches, most of them inevitably use information from other modalities or utilize the priori of body structure provided by SMPL to compensate for unstable results on sparse or noisy LiDAR point clouds. This can cause an increase in the complexity and latency of the system. Therefore, we think that an optimization-free approach based only on single-frame LiDAR is more practical. This leads us to learn a stable representation of joint for LiDAR-based 3D HPE which increases the reliability of the results and avoids cumbersome post-processing.

\subsection{Pre-training for human pose estimation}
Because pre-training allows models to learn generalized and dataset-independent representations, it has become an important technique to improve the performance of downstream tasks \cite{devlinBERTPretrainingDeep2019,heMomentumContrastUnsupervised2020,wangUnsupervisedPointCloud2021,liOnlineKnowledgeDistillation2021,liUniformMaskingEnabling2022}. On one hand, there have been attempts to use existing data for augmented representation learning. \cite{xuViTPose2022,anSHaRPoseSparseHighResolution2024} enhances the accuracy of Vision Transformer-based 2D pose estimation by masked self-supervised pre-training \cite{heMaskedAutoencoders2022}. \cite{qiuWeaklysupervisedPretraining3D2023} utilizes physical constraints provided by computational photography to conduct weakly supervised pre-training of 3D poses and improve the model's generalization. \cite{shanPSTMOPretrainedSpatial2022} brings an accuracy increase to 2D-to-3D lifting methods by applying a spatial-temporal mask of the skeleton. 

On the other hand, some methods try to use synthetic data to make up for the lack of labeled pose data. \cite{linMPTMeshPreTraining2024} utilizes the multi-camera projection of SMPL models to generate enough labeled samples for pre-training. \cite{weng3DHumanKeypoints2023a} proposes a LiDAR scene generation approach based on human mesh under ray casting. \cite{renLiDARaidInertialPoser2023,renLiveHPSLiDARbasedScenelevel2024a,renLiveHPSRobustCoherent2024} craft large-scale synthetic datasets based on \cite{mahmoodAMASSArchiveMotion2019} to obtain rich human priors. However, this method is limited by not considering the relative positions of bodies and LiDAR sensors and ignoring the diversity of occlusions. In this paper, we perform a richer and more reasonable sampling of spatial positions so that the model can learn point cloud representations from more aspects. Moreover, we construct more complex occlusion and noise point clouds to simulate the real environment.
\section{Methodology}
Our approach consists of two parts: a) a density-aware pose transformer (DAPT) for stable joint representation learning, b) a comprehensive LiDAR human synthesis \& augmentation for model pre-training. We adopt a two-stage training scheme, as shown in Fig.\ref{fig:framework}, we first pre-train the model on synthetic samples, which are generated by ray casting with random augmentations and occlusions. When the samples are fed into the proposed model, they are first encoded as sparse point features, Then, the valid information is progressively extracted to a set of joint anchors through multi-density exchange (MDE) modules and decoded into 1-D heatmaps. Finally, we input real samples into the model for fine-tuning. Note that the network architecture is shared between pre-training and fine-tuning with only slight differences in the loss functions.

In this section, we present the proposed method module by module.

\subsection{LiDAR human synthesis}
The quality and diversity of samples used for pre-training will directly affect the effect of downstream fine-tuning. Inspired by \cite{weng3DHumanKeypoints2023a}, we propose a more comprehensive strategy to sample and augment the scene.
\subsubsection{Scene sampling}
Given shape parameters $\vec{\beta}\in\mathbb{R}^{10}$ and pose parameters $\vec{\theta}\in\mathbb{R}^{72}$ sampled from a real-world captured SMPL database \cite{liLiDARCapLongrangeMarkerless2022}, a human instance is simply generated by the SMPL model:
\begin{equation}
    \{\mathbf{M}_h,\hat{\mathbf{J}}\}=\operatorname{SMPL}(\vec{\beta},\vec{\theta}),
\end{equation}
where $\mathbf{M}_h=\{\mathbf{V}_h\in{\mathbb{R}^{N_V\times3}},\mathbf{F}_h\in{\mathbb{Z}^{N_F\times3}}\}$ is a human mesh with $N_V$ vertices $\mathbf{V}_h$ and $N_F$ triangle faces $\mathbf{F}_h$. $\hat{\mathbf{J}}\in\mathbb{R}^{K\times3}$ are human joints. 

To simulate real ground, a ground mesh $\mathbf{M}_g=\{\mathbf{V}_g,\mathbf{F}_g\}$ is generated by a random normal vector $\vec{n}$ with size $s_g$, locating at the point with a minimum value of Z axis in $\mathbf{V}_h$.

Then, we randomly sample a polar coordinate with distance $r\in[4m,20m]$ and azimuth $\theta\in[-\pi,\pi]$, and convert it to Cartesian transition $\vec{t}$. Finally, the transformation is applied to $\mathbf{M}_h$ and $\mathbf{M}_g$ to get the scene mesh:
\begin{equation}
\begin{aligned}
\mathbf{M}=\{\mathbf{V}=(\mathbf{V}_h\cup\mathbf{V}_g)+\vec{t},\mathbf{F}=\mathbf{F}_h\cup\mathbf{F}_g\}.
\end{aligned}
\end{equation}
\subsubsection{Ray casting}
The LiDAR sensor obtains depth information through $360^\circ$ scans at different elevation angles. Thus, the 3D points it captured can be represented by polar coordinates $(r,\theta_l,\delta_l)$ within a laser grid $\mathcal{G}=\{\theta_i\}_{i=1}^{N_\theta} \times \{\delta_j\}_{j=1}^{N_\delta}$.
Since the synthetic mesh only occupies a small angle range, for faster sample generation and to facilitate the subsequent application of laser-level masks, we only intercept the laser within valid azimuth and elevation angles:
\begin{equation}\label{eq:effective_area}
\begin{aligned}
    \boldsymbol{\uptheta}&=\left[\theta_{min}(\mathbf{V}),\theta_{max}(\mathbf{V})\right]\cap\{\theta_i\}_{i=1}^{N_\theta},\\
    \boldsymbol{\updelta}&=\left[\updelta_{min}(\mathbf{V}),\updelta_{max}(\mathbf{V})\right]\cap\{\delta_j\}_{j=1}^{N_\delta},\\
    \mathbf{R}&=\{(\theta_{l},\delta_{l})|\theta_{l}\in\boldsymbol{\uptheta},\delta_{l}\in\boldsymbol{\updelta}\},
\end{aligned}
\end{equation}
where $\theta_{min}(\cdot),\theta_{max}(\cdot)$ take the minimum and maximum azimuth angles of the vertices, and for the elevation angles, the corresponding values are taken by $\delta_{min}(\cdot),\delta_{max}(\cdot)$. Then, ray casting is performed with the mesh to get the point cloud $\mathbf{P}\in\mathbb{R}^{N\times3}$ and which faces the lasers hit $\mathbf{H}\in\mathbb{Z}^{N}$:
\begin{equation}\label{eq:raycast}
    (\mathbf{P},\mathbf{H})=\operatorname{RayCast}(\mathbf{M};\mathbf{R}).
\end{equation}
Note that a label map describing the joints to which triangle faces belong is given by the SMLP model. Hence, the ground truth segmentation $\hat{\mathbf{S}}\in[0,1]^{N\times(K+1)}$ of $\mathbf{P}$ can be calculated by $\mathbf{H}$, where $\hat{\mathbf{S}}_{i,j}=1$ means the laser point $p_i\in\mathbf{P}$ belongs to the $j$ th joint. 
\subsubsection{Laser-level masking} 
To better simulate the occlusion in the real environment, we divide the effective LiDAR laser grid obtained in Eq.\ref{eq:effective_area} into patches with an empirical size $s_p=\left\lfloor{\min\{\max\boldsymbol{\uptheta}-\min\boldsymbol{\uptheta},\max\boldsymbol{\updelta}-\min\boldsymbol{\updelta}\}/8}\right\rfloor$. Then, $1-r_{keep}$ ratio of the total patches are being masked, which are denoted as $\mathcal{M}$. Finally, the point cloud is filtered by:
\begin{equation}
    \mathbf{P}_{syn}=\left\{p_i|p_i\in\mathbf{P},r_i\notin\mathcal{M}\right\},
\end{equation}
where $r_i\in \mathbf{R}$ is the ray producing hit point $p_i$.
\subsection{Density-aware pose transformer}
We present a density-aware pose transformer based on UNet-like point transformers \cite{wuPointTransformerV32024} to obtain stable joint representations of different densities. Specifically, the point cloud is firstly encoded into sparse point cloud features, which are subsequently fed into a decoder to be recovered to point-wise features. During the decoding process, we utilize an MDE module to export valid information from different pooling hierarchies and use 1D heatmaps to represent point locations, since heatmaps are generally more amenable to neural networks than coordinate regression.

\subsubsection{Point cloud feature deduction with MDE}
Given an input point cloud $\mathbf{P}\in\mathbb{R}^{N_0\times3}$, it was encoded into latent point features $f_M$ by an encoder $\mathcal{E}:\mathbb{R}^{N_0\times3}\rightarrow\mathbb{R}^{N_M\times D_M}$. Where $N_0$ is the initial point numbers, $N_M, D_M$ are the pooled point numbers and pooled point feature dimensions of $M$ th pooling level. 

As shown in Fig.\ref{fig:framework} right, it can be seen that at deeper pooling levels, the pooled point cloud shows a sparser spatial distribution, which inspires us to model joint-related features at these levels. Therefore, a set of learnable joint anchors $\mathbf{A}_M\in\mathbb{R}^{K\times D_M}$ is introduced. As the spatial dimensions of the features are expanded through the point transformer blocks, MDE progressively exchanges information between point features $f_m$ and joint anchors. Then, it updates the joint features by:
\begin{equation}
    \mathbf{A}_{m-1}=\operatorname{MDE}_m(\mathbf{A}_{m},f_m), m=M,M-1,\cdots,1.
\end{equation}
The MDE module first aligns the dimensions of joint features $\mathbf{A}_{m}$ to the dimensions of $f_m$ through an MLP, then applies self-attention on it while shortcutting it with padded $f_m$ for cross-attention.

\subsubsection{Heatmap decoder for joints} Two 1D heatmaps have been proven to be effective in representing 2D keypoint coordinates \cite{liSimCCSimpleCoordinate2022}. Inspired by this, we extend it to the representation of 3D coordinates of joints. Given the range and number of bins $N_{\{x,y,z\}}$, the heatmaps $h_{\{x,y,z\};i}$ of $i$ th joint can be predicted by MLPs on the corresponding axis. To get the target coordinate, we decode the 1D heatmaps by taking the peak location of $h_{\{x,y,z\}; i}$, and map them back to the coordinate through the counted range and bins.

\begin{figure*}[t]
  \centering
  \includegraphics[width=0.9\textwidth]{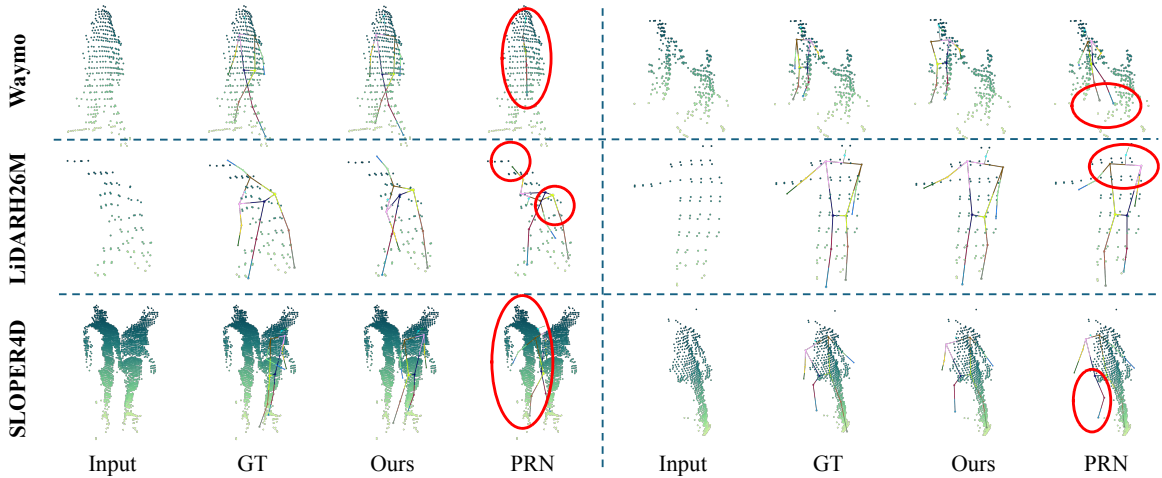}
  \caption{Visualization results on common datasets with challenging samples. Our method demonstrates strong stability on low-quality point clouds including occlusion, noise, and sparsity.}
  \label{fig:vis}
\end{figure*}

\begin{table*}[htbp]
    \centering
    \scalebox{0.88}{
    \begin{tabular}{l|c@{\quad}c@{\quad}c|c@{\quad}c@{\quad}c|c@{\quad}c@{\quad}c@{\quad}c|c@{\quad}c@{\quad}c@{\quad}c}
    \toprule
    Dataset & \multicolumn{3}{c|}{\textbf{Waymo}} & \multicolumn{3}{c|}{\textbf{HumanM3}}&\multicolumn{4}{c|}{\textbf{LiDARHuman26M}}&\multicolumn{4}{c}{\textbf{SLOPER4D}}\\ 
    Metrics & MPJPE & PCK3  & PCK5  & MPJPE & PCK3  & PCK5  & MPJPE & PA-   & PCK3  & PCK5  & MPJPE & PA-   & PCK3  & PCK5  \\ \midrule
    LidarCap &   -   &   -   &   -   &   -   &   -   &   -   & 79.31 & 66.72 & 86.00 & 95.00 &101.89 & 78.93 & 78.15 & 89.77 \\
    NE       &   -   &   -   &   -   &   -   &   -   &   -   & \textbf{72.23} & 61.67 & 87.94 & 95.79 & 96.80 & 76.70 & 79.22 & 90.51 \\
    LPFormer & 61.60 & 94.52 & 98.04 & 83.69 & 89.54 & 96.85 & 95.72 & 79.03 & 84.38 & 94.87 & 49.31 & 38.92 & 97.22 & 99.44 \\
    PRN      & 68.48 & 93.60 & 97.87 & 70.56 & 93.18 & 97.80 & 80.01 & 63.67 & 88.77 & 96.54 & 48.76 & 39.42 & 97.38 & 99.45 \\
    \rowcolor{gray!25}
    DAPT(ours)     & \textbf{51.59} & \textbf{97.34} & \textbf{98.98} & \textbf{59.76} & \textbf{95.39} & \textbf{98.30} & 73.47 & \textbf{58.40} & \textbf{90.54} & \textbf{97.06} & \textbf{28.01} & \textbf{21.52} & \textbf{99.30} & \textbf{99.87} \\ \bottomrule
    \end{tabular}
    }
    \caption{Comparison with other methods on IMU-annotated LidarHuman26M, SLOPER4D, and manually annotated Waymo Open Dataset, HumanM3. PA- stands for PA-MPJPE, and the units of MPJPE and PA-MPJPE are $mm$.The best results are highlighted in bold. For the same dataset, the skeleton structures are aligned to ensure fair comparison.} 
    \label{tab:result}
\end{table*}

\subsubsection{Pre-training}
The goal of pre-training is to allow the model to obtain human priors in the LiDAR-captured point clouds and understand the structure of the human body. Therefore, at this stage, to avoid learning preferences caused by differences in supervision intensity, following \cite{weng3DHumanKeypoints2023}, we still use coordinate regression on joint features and segmentation on point features for supervision. The joint regression loss $\mathcal{L}_{reg}$ is defined by:
\begin{equation}
    \mathcal{L}_{reg}=\sum_{i=1}^{K}\left\|\mathbf{J}_i-\hat{\mathbf{J}}_i\right\|_2\cdot v_i/\sum_{i=1}^{K}v_i,
\end{equation}
where $\mathbf{J}$ are the decoded joints coordinate from $\mathbf{A}_0$ through a shared decoder $\mathcal{D}_{reg}:\mathbb{R}^{D_0}\rightarrow\mathbb{R}^{3}$, and $\hat{\mathbf{J}}$ are the corresponding ground truth, $v_i$ is the joint visibility.
The segmentation loss $\mathcal{L}_{seg}$ is defined by:
\begin{equation}
    \mathcal{L}_{seg}=-\sum_{i=1}^{N_0}\sum_{j=1}^{K+1}\mathbf{S}_{i,j}\log(\hat{\mathbf{S}}_{i,j}),
\end{equation}
where $\mathbf{S}$ are the decoded part segmentation of points from $f_0$ through a shared decoder $\mathcal{D}_{seg}:\mathbb{R}^{D_0}\rightarrow\mathbb{R}^{K+1}$. Overall, we input synthetic data and minimize:
\begin{equation}
    \mathcal{L}_\text{pre} =\lambda_{reg}\mathcal{L}_{reg}+\lambda_{seg}\mathcal{L}_{seg},
\end{equation}
where $\lambda_{reg},\lambda_{seg}$ are loss weights.
\subsubsection{Fine-tuning}
The goal of the fine-tuning phase is to adapt the model to the joint annotations of different datasets and their specific distribution. Therefore, in this stage, we only enable the heatmap loss. We input real data and minimize:
\begin{equation}
    \mathcal{L}_\text{ft} = \mathcal{L}_{hm}=\sum_{c\in\{x,y,z\}}\sum_{j=1}^{K+1}D_{\text{KL}}(h_{c;j} \| \hat{h}_{c;j}),
\end{equation}
where $D_{\text{KL}}$ is the Kullback–Leibler divergence between GT and predicted 1D heatmaps.
\section{Experiment}

\subsection{Implementation Details}
For the model specification, we follow the typical configuration of PTv3 \cite{wuPointTransformerV32024}, with a voxelization grid size of 0.01. For data synthesis, the SMPL models are sampled from \cite{liLiDARCapLongrangeMarkerless2022}, and a simulated LiDAR sensor with 64 lines and 2650 angles is applied to perform ray casting, the $r_{keep}$ is set to 0.6 for laser-level masking. For model training, we perform 50 epochs in both pre-training and fine-tuning with AdamW \cite{loshchilovDecoupledWeightDecay2019} optimizer on 2 RTX 4090. We set the batch size to 64 and apply the cosine annealing decay strategy. For pre-training, we set the learning rate to $3\times10^{-4}$ and set $\lambda_{reg}=0.5$, $\lambda_{seg}=1.0$. For fine-tuning, the learning rate is set to $5\times10^{-4}$.
\subsection{Datasets}
We use four datasets with different scenarios and annotation methods to evaluate our method:

\noindent\textbf{LiDARHuman26M} \cite{liLiDARCapLongrangeMarkerless2022} A multi-modal dataset uses inertial measurement units (IMUs) captured human poses in SMPL format. It uses a fixed LiDAR sensor to capture a variety of daily actions within a range of 12m to 24m. The scenes are clear and ideal.

\noindent\textbf{SLOPER4D} \cite{daiSLOPER4DSceneAwareDataset2023} An IMU annotated dataset captured within a more realistic environment. A mobile LiDAR sensor is utilized to track a walking person for capturing point cloud data. Since the official train-test split is unavailable, we utilize the same data split following \cite{zhangNeighborhoodEnhanced3DHuman2024}.

\noindent\textbf{HumanM3} \cite{fanHumanM3MultiviewMultimodal2023} A multi-person pose dataset utilizing automatic annotation and manual review for accurate ground truth poses. It mainly includes multi-person scenes on sports fields, with point clouds and images captured by multiple sets of RGB-LiDAR units. Due to its large size and the small differences between adjacent frames, we only use 20\% of the data for training.

\noindent\textbf{Waymo Open Dataset v2} \cite{meiWaymoOpenDataset2022} A large-scale multi-task autonomous driving dataset with manual 3D pose annotations, contains 10K human instances.
\subsection{Metrics}
According to common practices, to evaluate model performance, we report \textbf{MPJPE}$\downarrow$ (Mean Per Joint Position Error), \textbf{PA-MPJPE}$\downarrow$ (Procrustes-Aligned Mean Per Joint Position Error), \textbf{PCK-3}$\uparrow$ (Percentage of Correct Keypoints with distance to GT lower than 30\% of torsal length), \textbf{PCK-5}$\uparrow$ (Percentage of Correct Keypoints with distance to GT lower than 50\% of torsal length). Note that we do not evaluate PA-MPJPE on Waymo and HumanM3, as their visibility labels of joints will interfere with the rigid body alignment process.

\subsection{Comparison methods} 
We compare our method with several state-of-the-art approaches. Specifically, we first evaluate against advanced Transformer-based models, LPFormer \cite{yeLPFormerLiDARPose2023} and PRN \cite{fanLiDARHMR3DHuman2023}, both of which are capable of inferring complete human body poses from single-frame LiDAR data. Additionally, we compare our method with LiDARCap \cite{liLiDARCapLongrangeMarkerless2022}, which leverages SMPL optimization and temporal information. We also include the Neighbor Enhanced (NE) 3HPE \cite{zhangNeighborhoodEnhanced3DHuman2024}, which incorporates information from the surrounding scene. To ensure a consistent comparison, we replace the point cloud backbone in PRN with PTv3. We note that LiDARCap and NE require consecutive frames and ground truth SMPL params, this limits the evaluation process on Waymo and HumanM3 datasets. Therefore, the results are not included. 

\begin{figure*}[t]
  \centering
  \includegraphics[width=0.9\textwidth]{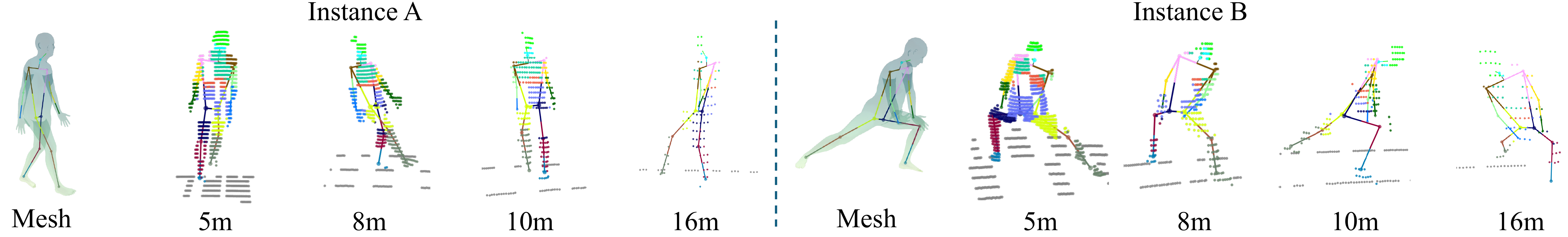}
  \caption{Visualization of synthesis samples. For the same SMPL human body, our method can produce diverse point clouds, and the laser level mask introduces extreme cases, which equips the model with
more robust human priors and the ability to recover human pose from low-quality point clouds.}
  \label{fig:vis_syn}
\end{figure*}

\begin{table}[t]
\centering
\scalebox{0.9}{
\begin{tabular}{l|c|c}
\toprule
Pre-training dataset & Waymo & LiDARH26M \\ \midrule
w/o & 59.2 & 76.5 \\
LIPD \cite{renLiDARaidInertialPoser2023} & 58.7 & 75.6 \\
FreeMotion \cite{renLiveHPSLiDARbasedScenelevel2024a} & 57.9 & 74.3 \\
NoiseMotion \cite{renLiveHPSRobustCoherent2024} & 59.0 & 76.3 \\
\rowcolor{gray!25}FreeMotion$\dagger$ & 54.6 & \textbf{73.2} \\
\rowcolor{gray!25}LiDARH26M$\dagger$ & \textbf{51.6} & 73.4 \\ \bottomrule
\end{tabular}
}
\caption{The effectiveness of our enhanced pre-training method, $\dagger$ means applying our data synthesis pipeline rather than the point cloud provided by the dataset.}
\label{tab:result_syn}
\end{table}

\begin{table}[t]
\centering
\small
\begin{tabular}{cccl}
\toprule
Anchor     & MDE        & Heatmaps   & MPJPE \small{(gain)} \\ \midrule
\multicolumn{3}{c}{PRN Baseline}         & 68.5  \\
\checkmark &            &            & 54.3 \small{(-14.2)}  \\
\checkmark & \checkmark &            & 52.3 \small{(-16.2)}  \\
\checkmark &            & \checkmark & 53.8 \small{(-14.7)}  \\
\rowcolor{gray!25}
\checkmark & \checkmark & \checkmark & \textbf{51.7} \small{(-16.8)}  \\ \bottomrule
\end{tabular}
\caption{Ablation study of progressively enabling our proposed components.}
\label{tab:abl_components}
\end{table}

\begin{table}[t]
  \centering
  \begin{subtable}[t]{0.55\linewidth}
  \centering
  \scalebox{1.0}{
    \setlength{\tabcolsep}{1.8mm}
      \begin{tabular}{cl}
        \toprule
        Method                 & MPJPE \small{(gain)} \\ \midrule
        scratch                & 59.2 \\
        +pre-training          & 52.9 (-\small{6.3}) \\
        +scene sampling        & 52.4 (-\small{6.8})\\
        \rowcolor{gray!25}
        +laser masking         & \textbf{51.7} (-\small{7.5})\\ \bottomrule
      \end{tabular}
    }
    \subcaption{Influnce of augmentations}
    \label{tab:abl_aug}
  \end{subtable}
  \begin{subtable}[t]{0.4\linewidth}
    \centering
    \scalebox{0.85}{
    \setlength{\tabcolsep}{1.8mm}
        \begin{tabular}{cc}
        \toprule
        $r_{keep}$ & MPJPE \\ \midrule
        0.5        &53.0\\
        \rowcolor{gray!25}
        0.6        &\textbf{51.7}\\
        0.7        &51.9\\
        0.8        &52.5\\
        0.9        &52.4\\ \bottomrule
        \end{tabular}
    }
    \subcaption{Influnce of $r_{keep}$}
    \label{tab:abl_rkeep}
  \end{subtable}
  \caption{Ablation study of pre-training related components.}
  \label{tab:abl_pretrain}
\end{table}

\begin{figure}[ht]
\begin{subfigure}{.49\linewidth}
  \centering
  \includegraphics[width=\linewidth]{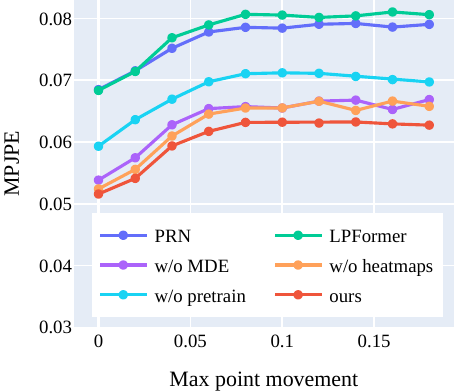}
  \caption{Point jittering}
  \label{fig:stb_jit}
\end{subfigure}
\hfil
\begin{subfigure}{.49\linewidth}
  \centering
  \includegraphics[width=\linewidth]{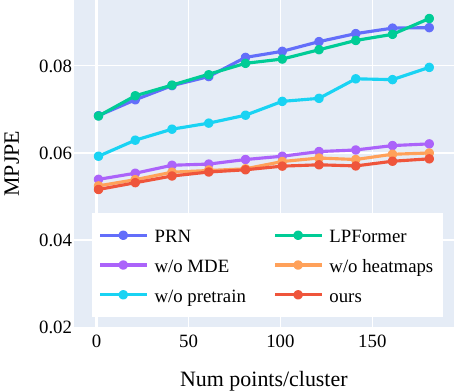}
  \caption{Noise point clusters}
  \label{fig:stb_gpcn}
\end{subfigure}
\caption{Stability evaluation of our method. We observe that the performance degradation of the proposed method is within a more acceptable range.}
\label{fig:stb}
\end{figure}

\begin{figure}[t]
  \centering
  \includegraphics[width=0.9\linewidth]{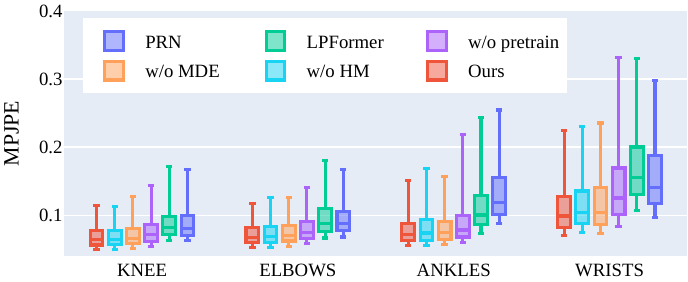}
  \caption{Errors comparison of the most challenging human end joints. The worst 600 instances of each method are included in the statistics.}
  \label{fig:joint_statistic}
\end{figure}

\subsection{Quantitative Results}
Tab.\ref{tab:result} presents the comparison results, where our method consistently achieves outstanding results across four datasets. On LiDARHuman26M with clear scenes, our method obtains comparable MPJPE to the current best NE but with a reduced PA-MPJPE ($-3.2mm$), suggesting closer alignment with GT. On the more challenging SLOPER4D, our method gets a remarkable $28.01mm$ MPJPE, representing a significant improvement over the current best PRN ($-20.7mm$). In the motion-focused dataset HumanM3, our method records an MPJPE of $59.76mm$, surpassing PRN by $10.8mm$. Finally, on the Waymo dataset, which centers on autonomous driving scenarios, our method achieves an MPJPE of $51.59mm$, improving by $10.0mm$ over the SOTA method LPFormer. Overall, our method consistently achieves higher PCK-30 and PCK-50 scores, indicating greater robustness and value of practical usage.

\subsection{Statbility Evaluation}
To evaluate the stability of our method, we introduce various types of disturbance to the point cloud and re-evaluate the model's performance. Specifically, we apply clusters of noise points at different locations and introduce positional offsets to each point. Results are presented in Fig.\ref{fig:stb}.
\subsubsection{Point jittering}
We add point noise to each point coordinate of the input and use different thresholds for clipping. The results are shown in Fig.\ref{fig:stb_jit}. It can be observed that compared with baseline methods, our method has less performance degradation, and the proposed MDE module also plays a positive role in dealing with disturbances.
\subsubsection{Noise clusters}
We simulate extreme input conditions by adding clusters containing varying amounts of noise points to the input point cloud. As shown in Fig.\ref{fig:stb_gpcn}, as the number of noise points per cluster increases, our method yields more stable results. Notably, our pre-training substantially enhances the model's ability to handle noisy point clouds.
\subsubsection{Errors on end joints}
We evaluate the errors on the most challenging end joints, and the results are shown in Fig.\ref{fig:joint_statistic}. Our method demonstrates smaller average error and variance on these joints. Especially, the proposed pre-training method effectively stabilizes the prediction of ankles and wrists.
\subsection{Ablation study}
\subsubsection{Model components}
In Tab.\ref{tab:abl_components}, we conduct ablation studies on the Waymo dataset to evaluate the contributions of the key components by incrementally enabling each proposed module. Specifically, row 1 represents the PRN baseline. Row 2 shows the results after introducing joint anchors and a 4-layer Transformer on top of the off-the-shelf PTv3. Row 3 demonstrates the effect of enabling the proposed MDE module on multi-scale point cloud features. In row 4, the coordinate-based decoding is replaced with heatmap-based decoding. The final row presents the results with all modules enabled. The combined effect of joint anchors and the MDE module leads to a substantial $16.2mm$ improvement in MPJPE over the baseline. Replacing the decoder with a heatmap-based one results in an additional $0.6mm$ improvement.

\subsubsection{Pre-Training}
In Tab.\ref{tab:abl_pretrain}, we conduct ablation studies to assess the impact of our proposed pre-training strategies. Tab.\ref{tab:abl_aug} evaluates the effectiveness of the enhancements: the first row represents training without any pre-training, the second row uses the same pre-training strategy as \cite{weng3DHumanKeypoints2023}, the third row implements scene resampling, and the fourth row further incorporates the proposed laser level masking. Comparing rows 1 and 2, the inclusion of pre-training brings an MPJPE reduction of $6.3mm$. Further comparison of rows 1 and 4 shows that incorporating both scene resampling and laser-level masking leads to an even greater improvement of $7.5mm$. Moreover, as shown in Tab.\ref{tab:abl_rkeep}, to strike a balance between providing sufficient information for human pose reconstruction and enabling the model to recover human structures from occluded point clouds, we vary the proportion of the laser level mask $r_{keep}$ from $0.5$ to $0.9$ to determine the optimal parameter. In addition, as shown in Tab.\ref{tab:result_syn}, we also compare our synthesis pipeline with LIPD \cite{renLiDARaidInertialPoser2023}, FreeMotion \cite{renLiveHPSLiDARbasedScenelevel2024a}, and NoiseMotion \cite{renLiveHPSRobustCoherent2024}, which also use synthetic data. Pre-training with data synthesized by the proposed method can bring more performance improvements.

\subsection{Visualization}
\subsubsection{Prediction results}
Fig.\ref{fig:vis} presents the visualization results of our method. We select some challenging samples from the dataset, and our method consistently produces more stable results. Row 1 left shows a walking person from the Waymo dataset, but the left-right symmetry of the body is unclear. A segmentation-regression-based method produces wrong results, attempting to disregard the human body’s rigid structure to mitigate inference errors caused by left-right ambiguity. In contrast, our method penalizes such ambiguities in the heatmap, effectively avoiding this issue. Row 2 illustrates the detection results on sparse point clouds from the LiDARHuman26M dataset. In this scenario, our method effectively recognizes human body orientation from limited clues. Lastly, as depicted in row 1 right and row 3 left, when multiple human instances are unexpectedly introduced by the human detector, our approach does not confuse the two individuals. In summary, our method demonstrates strong capability in robust inference.
\subsubsection{Synthetic samples}
We present the synthesis samples of our approach in Fig.\ref{fig:vis_syn}. It is evident that scene resampling produces diverse synthetic samples, effectively generating both sparse and dense point clouds of human instances. Additionally, the laser-level masks introduce more complex occlusions, which in turn equip the model with more robust human priors and point cloud representations.

\section{Conclution}
We propose a novel method for robust LiDAR-based 3D human pose estimation, with two main contributions. First, we introduce a density-aware pose transformer, which employs joint anchors and special exchange modules to extract valid features from point clouds of varying densities, enabling explicit learning of stable keypoint representations. Second, we present a comprehensive LiDAR-based human synthesis and augmentation approach for model pre-training. By integrating more randomized position sampling, ground modeling, and laser-level masking, we generate highly realistic and challenging samples. Qualitative and quantitative evaluations demonstrate that with the synergy of the proposed model and pre-training strategies, our method achieves state-of-the-art performance across multiple datasets.

Overall, our approach provides a comprehensive solution for LiDAR-based 3D HPE. Future work can try to address left-right reversal and multi-frame jitter in timing or perform pose-based human behavior understanding tasks.

\section*{Acknowledgments}
This work was supported by the National Science Fund of China under Grant Nos.U24A20330, 62172222, 62361166670, and the National Key Research and Development Program of China (International Collaboration Special Project, No. SQ2023YFE0102775).
\bibliography{aaai25}

\end{document}